%% file: main.tex
\documentclass[onecolumn]{cohere}

\usepackage[utf8]{inputenc} 
\usepackage[T1]{fontenc}    
\usepackage{hyperref}
\usepackage{url}            
\usepackage{booktabs}       
\usepackage{amsfonts}       
\usepackage{nicefrac}       
\usepackage{microtype}      
\usepackage{xcolor}         
\usepackage{multirow}
\usepackage{graphicx}
\usepackage{enumitem} 
\usepackage{float} %
\usepackage{multirow}
\usepackage{makecell}
\usepackage{arydshln}
\usepackage{wrapfig}
\usepackage{nameref}
\usepackage{hyperref}
\usepackage[symbol]{footmisc}

\definecolor{orange}{rgb}{1.0, 0.6, 0.1}

\usepackage{listings}

\usepackage{xcolor}

\definecolor{codegreen}{rgb}{0,0.5,0}
\definecolor{codegray}{rgb}{0.5,0.5,0.5}
\definecolor{codepurple}{rgb}{0.68,0,0.72}
\definecolor{codered}{rgb}{0.8,0,0.2}
\definecolor{backcolour}{rgb}{0.98,0.98,0.98}

\title{
Pushing Mixture of Experts to the Limit: Extremely Parameter Efficient MoE for Instruction Tuning
}

\author{
    name={Ted Zadouri},
    affiliation={Cohere for AI},
    email={ted@cohere.com}
}
\author{
    name={Ahmet Üstün},
    affiliation={Cohere for AI},
    email={ahmet@cohere.com}
}
\author{
    name={Arash Ahmadian\textsuperscript{\dag}},
    affiliation={Cohere for AI},
    email={arash@cohere.com}
}
\author{
    name={Beyza Ermi\c{s}},
    affiliation={Cohere For AI},
    email={beyza@cohere.com}
}
\author{
    name={Acyr Locatelli},
    affiliation={Cohere},
    email={acyr@cohere.com}
}
\author{
    name={Sara Hooker},
    affiliation={Cohere for AI},
    email={sarahooker@cohere.com}
}

\date{\today}

\abstract{

The Mixture of Experts (MoE) is a widely known neural architecture where 
an ensemble of specialized sub-models optimizes overall performance with a constant computational cost. However, conventional MoEs pose challenges at scale due to the need to store all experts in memory. In this paper, we push MoE to the limit. We propose extremely parameter-efficient MoE by uniquely combining MoE architecture with lightweight experts.Our MoE architecture outperforms standard parameter-efficient fine-tuning (PEFT) methods and is on par with full fine-tuning by only updating the lightweight experts -- less than 1\% of an 11B parameters model. Furthermore, our method generalizes to unseen tasks as it does not depend on any prior task knowledge. Our research underscores the versatility of the mixture of experts architecture, showcasing its ability to deliver robust performance even when subjected to rigorous parameter constraints. Our code used in all the experiments is publicly available here: \url{https://github.com/for-ai/parameter-efficient-moe}.

}

\begin{document}
\footnotetext[0]{\textsuperscript{\dag}Also affiliated with the University of Toronto \& the Vector Institute for Artificial Intelligence.}

\section{Introduction}
\label{sec:intro}

\begin{figure*}[!htb]
    \centering
    \begin{subfigure}[t]{0.9\linewidth}
        \includegraphics[width=0.99\linewidth]{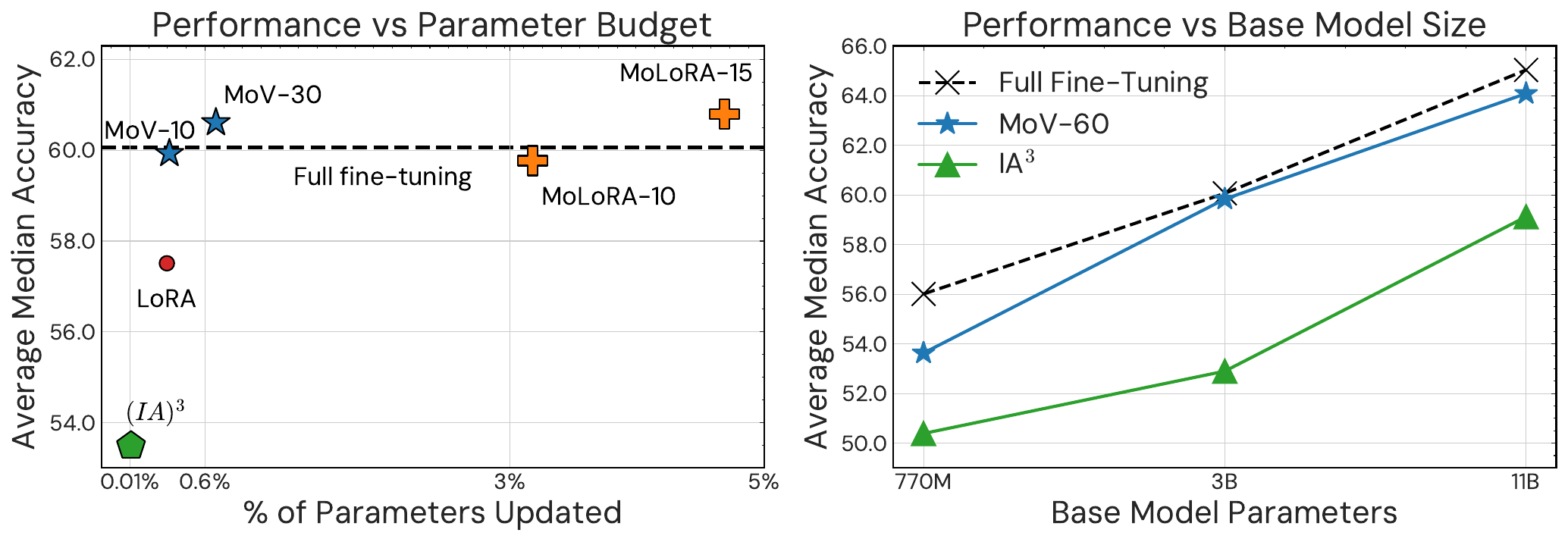}
    \end{subfigure}
        \caption{\emph{Left}: Our mixture of PEFT experts outperforms SOTA single PEFT methods using a comparable amount of parameters demonstrated for T5-XL (3B). \emph{Right}: Mixture of PEFT approach scales up to 11B; with tiny parameter updates, it approximates or matches full fine-tuning performance.}\label{figure:PGA_variance}
        \label{fig:model_params}
\end{figure*}

A conventional training paradigm is to apply the weights of a model to each input. Arguably, this is not efficient since a given input may not need all of a model's capacity. In contrast, MoEs build on the premise that sub-modular components -- so called experts -- can specialize to different types of inputs. This emphasis on conditional computation has important efficiency side-effects such as constant inference cost. This has made MoEs an area of significant research and widespread adoption in the era of large-scale Transformers where scaling has increased deployment and latency costs \citep{shazeer2018meshtensorflow,riquelme2021scaling,du2022glam,fedus2022switch}.

While the majority of work to-date has focused on MoEs as a pretraining strategy,the inherent motivation of MoEs is not confined solely to pretraining. In fact, the merits of MoEs are arguably well suited to an \textit{instruction fine-tuning} setting where the data is often deliberately structured to represent a diverse set of tasks, often referred to as multi-task finetuning \citep{chung2022scaling,wei2022finetuned,sanh2022multitask, longpre2023flan, muennighoff2023crosslingual}.

In this work, we pose the question \textit{can we leverage MoEs for instruction fine-tuning?} One of the main drawbacks of MoEs paradigm is that it introduces an extreme amount of total parameters \citep{fedus2022switch}. 
Despite the conditional computation, fully fine-tuning MoE architecture is extremely computationally demanding given the need to update all the parameters. For most practitioners, given the scale of modern LLMs \citep{brown2020language, touvron2023llama, kaplan2020scaling, anil2023palm} this is an infeasible computational cost.

Thus, we focus on a more realistic setting for everyday practitioners -- \textit{can we successfully apply MoEs to parameter-efficient fine-tuning (PEFT)} methods such as $\textrm{(IA)}^3$ \citep{liu2022fewshot} or LORA \citep{hu2021lora} which only fine-tune a far smaller number of parameters. This is a significant challenge not only since our aim is to update only a small percentage of all parameters but as we also navigate the optimization challenges inherent to MoEs already noted by prior work \citep{chen2022towards} in a more constrained environment.

In this work, we propose a new framework that leverages the benefits of MoE in a severely constrained computational environment. We introduce \textbf{Mixture of Vectors (MoV)} and \textbf{Mixture of LORA (MoLORA)}, a parameter-efficient adaptation of the Mixture of Experts approach. Unlike the standard MoE, our framework can be utilized in a parameter-limited setting due to its lightweight nature. Remarkably, our method achieves performance parity with full fine-tuning on unseen tasks by updating less than 1\% of the parameters. It also easily outperforms base parameter-efficient techniques like $\textrm{(IA)}^3$ or LORA.

We achieve consistent results across
T5 models \citep{raffel2020exploring} ranging from 770M to 11B across $12$ different tasks from 55 datasets P3 \citep{sanh2022multitask}. In summary, our contributions are as follows:
\begin{enumerate}[label=(\roman*)]
    \item We present extremely parameter-efficient MoEs. This architecture leverages MoEs in a more realistic setting using modular and lightweight experts. Our MoEs can be used to fine-tune a dense model by updating less than 1\% of its parameters. 
    \item Instruction fine-tuning with our proposed methods consistently outperforms traditional parameter efficient methods on unseen tasks, while maintaining high parameter efficiency across different scales. The mixture of $\textrm{(IA)}^3$ vectors (MoV) achieves up to 14.57\% and 8.39\% improvements over the standard $\textrm{(IA)}^3$ at 3B and 11B model sizes respectively. This superiority holds across different model sizes, types of experts and trainable parameter budgets.

    \item We show that our recipe can match the performance of \emph{full fine-tuning} at large scales while updating a tiny fraction of the model parameters. Our results across 8 unseen tasks show that our MoV which updates just 0.32\% and 0.86\% of the parameters in the 3B and 11B models achieves \emph{higly competitive} performance to full fine-tuning with a significantly reduced computational cost. 

    \item Finally, we present an extensive set of ablation studies that systematically evaluate the efficacy of various MoE architectures and PEFT strategies at various model sizes, different adapter types, the number of experts, routing mechanisms, and the importance of optimizing hyper-parameters, especially given the sensitivity of MoE. 
\end{enumerate}


\input{background.tex}

\input{methods.tex}

\input{experiments.tex}

\input{results}

\input{related_work.tex}

\section{Conclusion}
\label{sec:conc}

This work introduces MoEs in an extremely computationally limited environment. We propose introduce the Mixture of Vectors (MoV) and Mixture of
LoRA (MoLORA) to mitigate the challenges associated with scaling instruction-tuned LLMs at scale. Our method outperforms parameter-efficient techniques and achieves performance parity with full fine-tuning on unseen tasks by updating less than 1\% of the 3B and 11B model parameters.
This percentage may vary depending on the base model size and the number of experts involved. Our extensive experiments, including rigorous ablations across model sizes, representation of tokens vs embeddings, soft vs top-k routing, confirm the effectiveness of our approach across diverse unseen tasks, highlighting its superior accuracy and computational efficiency. Furthermore, our framework's versatility seamlessly integrates with other parameter-efficient strategies and remains compatible with efficiency-enhancing techniques such as quantization.

\textbf{Limitations} A primary constraint of our experimental framework is its focus on text-to-text models, such as T5, without extending the evaluation to decoder-only such as GPT style models. We leave this as the subject of future work. Additionally, our assessment is exclusively within the context of fine-tuning. Exploration of its efficacy during the pre-training phase remains an avenue for future research.

\bibliography{references}
\newpage
\input{appendix.tex}

\renewcommand{\thesection}{\Alph{section}}

\end{document}

%% file: background.tex
\section{Methodology}
\label{sec:bg}

The instruction tuning setup is formulated as such where there are set of tasks 
which are divided into training and held-out evaluation tasks,
$T = T_{\text{train}} \cup T_{\text{eval}}$. 
The base pretrained model is first fine-tuned on $T_{train}$ and then evaluated in a zero-shot manner on each unseen task from $T_{eval}$. 
The standard approach is fine-tuning all model parameters that cause high compute and memory costs. Our method offers an efficient alternative using parameter-efficient mixture of experts. In this section, we describe our framework in detail. 

\subsection{Parameter-efficient Fine-tuning with $\textrm{(IA)}^3$ and LORA Adapters}

In this work, we push the mixture of expert (MoE) architecture to an extreme degree of parameter efficiency using \textit{parameter-efficient fine-tuning} (PEFT) methods. PEFT methods address the challenges associated with updating a large number of parameters -- especially emerging at scale when fully fine-tuning an LLM -- by restricting weight updates to a limited number of parameters.  To show how our method scales with different PEFT techniques, we experiment with both $\textrm{(IA)}^3$ and LORA. These methods add a small number of parameters to the existing pre-trained model. We briefly introduce each PEFT method below:

\textbf{$\textrm{(IA)}^3$} introduces three new vectors, $l_\text{k} \in \mathbb{R}^{d_\text{k}}$, $l_\text{v} \in \mathbb{R}^{d_\text{v}}$, $l_\text{ff} \in \mathbb{R}^{d_\text{ff}}$ which re-scale key and value activations in self-attention, and intermediate activations in position-wise feed-forward layers:

\begin{align*}
 \text{softmax}\left(\frac{Q(l_\text{k} \odot K^T)}{\sqrt{d_\text{k}}}\right)(l_\text{v} \odot V); ~~(l_\text{ff} \odot \gamma~(W_1x))W_2 \tag{$\textrm{(IA)}^3$}
\end{align*}

where $Q$, $K$, $V$ are query, key, and value projection matrices for self-attention, and $W_1$, $W_2$ are frozen weights of the feed-forward layers in the pretrained model. Since $\textrm{(IA)}^3$ only updates $l_\text{k}$, $l_\text{v}$, $l_\text{ff}$ rescaling vectors for each Transformer layer\footnote{For an encoder-decoder model with L number of layers in both sides, $\textrm{(IA)}^3$ only introduces $L(d_\text{k}+d_\text{v}+d_\text{ff})$ new parameters for encoder and $L(2d_\text{k}+2d_\text{v}+d_\text{ff})$ for decoder, due to the additional encoder-decoder attention block.}, it is extremely parameter-efficient. For the 3 billion parameter T5 model \citep{raffel2020exploring}, it only updates $0.018\%$ of the total parameters.  

Note that, unlike adapters \citep{houlsby2019parameterefficient} or prompt-tuning \citep{lester2021power}, the number of new parameters inserted by $\textrm{(IA)}^3$ is determined by the architecture as the scaling vectors need to be the same size with the corresponding activation dimensions. 

\begin{figure}[t]
\begin{minipage}{0.50\textwidth}
\centering
    \includegraphics[width=0.8\linewidth]{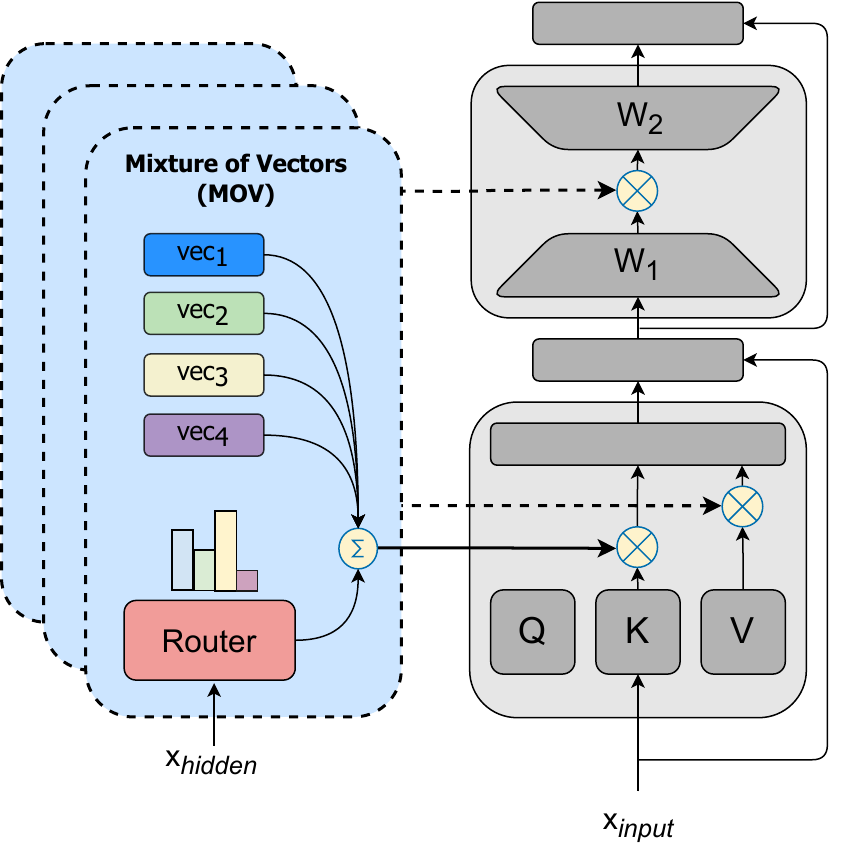}
    \label{fig:mov-arch}  
\end{minipage}
\hspace{2pt}
\begin{minipage}{0.56\textwidth}
\newsavebox{\algobox}
\begin{lrbox}{\algobox}%

\begin{lstlisting}[language=Python]
class MOV_Layer(nn.module):
  n_experts: int # number of experts

  def call(self, inputs):
    # inputs shape: [batch, seq, h_dim]
    batch, seq, h_dim = inputs.shape

    # MOV scaling vectors: [n_experts, h_dim]
    mov_vectors = self.param('mov_scalers',
      nn.init.ones(), (self.n_experts, h_dim))

    # router probs: [batch, seq, n_experts]
    router_probs = self.router(inputs, 
      self.n_experts, dtype='float32')

    # combined vector: [batch, seq, h_dim]
    mov_combined = jnp.einsum('...e,ed->...d', 
                              router_probs, 
                              mov_vectors)

    return inputs * mov_combined
\end{lstlisting}
\end{lrbox}
\scalebox{0.85}{\usebox{\algobox}}
\label{fig:mov-algo}  
\end{minipage}
\vspace{1pt}
\caption{\emph{Left}: Overview of the MoV architecture highlighting soft-merging where only the vectors and router are updated for each multi-head attention block, as denoted by color. \emph{Right}: JAX-like pseudo-code illustrating the core implementation of a MoV layer.}
\label{fig:mov-architecture}
\end{figure}

\textbf{Low-Rank adaptation} \citep[LORA;][]{hu2021lora} optimizes low-rank decomposition of dense layers in LLMs. For a pre-trained weight matrix $W_0 \in \mathbb{R}^{d_\text{m}\times d_\text{p}}$ and input activation $x \in \mathbb{R}^{d_\text{m}}$, LORA decomposes $W_0$ into two low-rank matrices: 

\begin{align*}
    h = W_0  + \Delta W_x = W_0 +  BAx \tag{LORA}
\end{align*}

where $B \in \mathbb{R}^{d_\text{p} \times r}$ $A \in \mathbb{R}^{r \times d_\text{m}}$, and the rank $r = \min(d_\text{m},d_\text{p})$. During fine-tuning, all pretrained weights are frozen, and only $A$ and $B$ matrices are updated.

LORA adaptation can be used for all the linear layers in each Transformer block including query $Q$, key $K$, value $V$, and output $O$ of the self-attention and the feed-forward layers $W_1$ and $W_2$. Unlike $\textrm{(IA)}^3$, LORA adaptation offers more flexibility in terms of the parameters used. We can adjust the capacity by incrementing the rank $r$ of the matrix decomposition until it reaches its maximum, determined by $r = \min(d_\text{m},d_\text{p})$. To illustrate its parameter efficiency, for a T5 3B model, LORA with a rank of 4, updates $0.3\%$ of the model parameters.

%% file: methods.tex
\subsection{Extremely Parameter Efficient Mixture of Experts}
\label{sec:methods}

We propose an extremely parameter-efficient Mixture of Experts (MoE) framework that leverages lightweight ``adapters'' as experts on top of a pretrained dense model. Concretely, the MoE is a family of neural network architecture that enables conditional computation through multiple experts that are activated based on a gating mechanism (router). An MoE layer consists of a router network $R$ and a set of $n$ experts $E_1, ..., E_n$ where each expert $E_i$ is a parameterized function. Following \citet{fedus2022switch}, our router network commonly consists of a dense layer with trainable weights $W_g \in \mathbb{R}^{d_\text{m} \times n}$ followed by a \textit{softmax} function which takes an intermediate token representation $x$ as input and combines the output of each expert based on the gating scores $s_1, ..., s_n$:

\begin{gather*}
 s_i = R(x)_i = \text{softmax}(W^T_g x) \tag{Router}
 \\
 y = \sum_{i=1}^n s_i\cdot E_i(x) \tag{MoE} 
\end{gather*}

For Transformer models \citep{vaswani2023attention}, dense feed-forward layers are replaced by MoE layers where each expert $E_i$ corresponds to an independent dense feed-forward network. This multiplies the total number of model parameters as each expert size and number of experts increase. However, in our parameter-efficient MoE architecture, we replace each expert with a lightweight PEFT adapter such as $\text{(IA)}^3$ vectors or LORA adapters. During fine-tuning, pretrained weights of dense layers remain fixed, while experts and router layers are trained from scratch. Unlike the standard MoE, our lightweight experts learn to adapt the pretrained Transformer layers in the fine-tuning time. In this way, our MoE framework requires a limited number of parameter updates and does not introduce a huge model size in total.  

In addition to parameter efficiency, our selection of PEFT adapters enables routing computation with \emph{soft merging}. 
Concretely, since both $\text{(IA)}^3$ vectors and LORA adapters are linear functions, we compute a weighted average of experts first and then apply a PEFT transformation using the combined expert $E_{mix}$ similar to \citet{muqeeth2023soft}:

\begin{gather*}
 E_{mix} = \sum_{i=1}^n s_i\cdot E_i; ~~ y =  E_{mix}(x)
 \tag{Soft Merging}
\end{gather*}

We call the variants of our method as \textit{Mixture of Vectors} (\textbf{MoV}) and \textit{Mixture of LORA} (\textbf{MoLORA}) that leverage $\text{(IA)}^3$ vectors or LORA adapters as experts respectively, both demonstrating consistent gains over the corresponding PEFT method. Figure \ref{fig:mov-architecture} shows the architecture of a MoV layer together with the corresponding pseudo-code. Only updating a small fraction of parameters through MoV and MoLORA has multiple practical benefits not only to training but to inference time, with the latter being unique to MoE architectures. We provide a brief overview of these gains below: 
 
\textbf{Efficiency in training} Our extremely parameter-efficient MoE formulation leads to a significant reduction in memory. The freezing of most parameters during training reduces the computational overhead of calculating gradients for model parameters but also reduces the memory requirements of storing the optimizer states for the model. The latter can be quite significant depending on the choice of the optimizer, for instance, variants of Adam \citep{kingma2017adam} including AdamW \citep{loshchilov2019decoupledadamw}, require twice the memory required for each parameter, to store the optimizer states (estimates for first and second moments) whereas Adafactor \citep{shazeer2018adafactor} reduces this overhead roughly by half through factored estimation of the second-order parameter moments. 
 
\textbf{Efficiency at inference} The inherent structural modularity of our MoV and MoLORA methods allows for significant memory gains at inference time. For traditional MoE models, many copies of the full-fledged feed-forward blocks (or even complete replicas of the model based on specific architecture) need to be stored in memory at inference time which is an expensive undertaking. With our methods, regardless of the exact type, only a single copy of the model backbone needs to be stored in memory in addition to lightweight parameter-efficient experts. This leads to a significant reduction in the memory requirements at inference time.

%% file: experiments.tex
\section{Experiments}
\label{sec:experiments}

\noindent \textbf{Dataset} ~ We conduct instruction-tuning experiments using a comprehensive set of prompt instructions from the Public Pool of Prompts (P3) dataset \cite{sanh2022multitask}. We follow the same procedure as \cite{raffel2020exploring} where each task is converted into the format provided templates in \citep{sanh2022multitask}. P3 is a collection of 62 datasets covering a wide variety of tasks. 

\noindent \textbf{Experimental Setup} ~ For the base pretrained models, we use T5 v1.1+LM adaptation \citep{lester2021power} that includes T5 models of different sizes ranging from 770M to 11B parameters. For all experiments, we fine-tune using Adafactor optimizer \citep{shazeer2018adafactor} with a learning rate of $3e^{-4}$. We set the sequence length to 1024 for the input and 256 for the target following to \citet{sanh2022multitask}. For all parameter-efficient MoE variants, we fine-tune T5 models using a batch size of 32 over 500K steps.

\noindent \textbf{Baselines} ~ We compare our mixture of parameter-efficient experts against both T0 baseline as the fully fine-tuned model, and the standard parameter-efficient fine-tuning methods $\text{(IA)}^3$ and LORA. For T0 baselines, based on our experiments with different hyperparameters, we find that a larger batch size and learning rate result in better performance, thus, we replicated T0 by fine-tuning for 10k steps with a batch size of 256, and a learning rate of $1e^{-3}$, 
following \citet{phang2023hypertuning} -- these hyperparameters achieve significantly higher results as shown in Table \ref{tab:main-results}. For $\text{(IA)}^3$ and LORA with rank=4, we use the same training hyper-parameters such as learning rate of $3e^{-4}$ and batch of 32 over 500k steps.

\noindent \textbf{Metrics} ~ Following the zero-shot evaluation presented in T0 \cite{sanh2022multitask}, we test our method and the baselines on 8 held-out (unseen during training) datasets -- ANLI \citep{nie2020adversarial}, HellaSwag \citep{zellers2019hellaswag}, WinoGrande \citep{sakaguchi2019winogrande}, and 5 Super Glue datasets \cite{wang2020superglue}. These datasets cover different tasks ranging from coreference resolution, natural language inference, multiple-choice question answering, story completion, and word sense disambiguation. We calculate the median accuracy for each evaluation dataset across different prompt templates and then report the per-dataset result together with an average across all datasets. We also include the mean accuracy for all evaluation datasets in the Appendix. 

\noindent \textbf{Infrastructure} ~ All experiments were conducted on TPU v4 machines up to 256 pod slices. For training, evaluation, and inference of all the models experimented, we used SeqIO and T5X \citep{roberts2022t5x} frameworks that enable data and model parallelization across TPU cores with integrated sequential data processing.

\subsection{Ablations}

 Given no work to-date has studied MoE in extremely parameter-efficient settings, we also seek to understand key characteristics of our proposed methdology by running rigorous ablations. We detail both briefly, along with the experimental set-up below:

\noindent \textbf{Routing Input: Token vs Sentence Embeddings} ~ \emph{How does a pronounced inductive bias for task representations in the form of instruction embedding affect routing and downstream generalization?} In our main MoV and MoLORA methods, router layers take intermediate embeddings of input tokens as input similar to other MoE architectures \citep{shazeer2017outrageously,fedus2022switch}. However, as an alternative, a sentence embedding can be computed for each instruction (prompt with corresponding input) and be used as input for the router \citep{ye2022eliciting}. To compare both -- sentence embeddings for each instruction were derived using the Sentence-T5 encoder \citep{ni2022sentence}, trained with the T5-XL retrieval model \citep{ni2021large}. This encoder was initialized from the pretrained T5 and trained on diverse data sources as outlined in \cite{ni2022sentence}. Without additional fine-tuning, each instruction sequence which consists of a prompt template and the input sentence, was passed to retrieve the embeddings with a dimension of $768$.

\noindent \textbf{Routing Strategy: Soft vs Discrete} ~ \emph{What is the best routing strategy in parameter-efficient MoEs?}
In our MoE framework, we use soft merging of experts as routing strategy. Soft merging refers to a weighted average of all the experts computed within a specified routing block. As an alternative, discrete top-k routing strategy as used in standard MoE architectures introduces the sparsity and decreases the amount of computation \citep{shazeer2018meshtensorflow,fedus2022switch}. In the top-k routing approach, rather than considering all experts for a decision, only the top 'k' experts, determined by the router, are chosen for the computation. 
Note that, although the computation is conditional to the top-k experts, the required memory depends on the total number of experts. 

We evaluate top-k selection with $k= \{1,2\}$ as they were proposed by previous work \citep{shazeer2017outrageously,fedus2022switch}. Results for these strategies are elaborated in Section~\ref{sec:token_v_embedding}. Additionally, we assess discrete routing with top-k using \textit{load balancing} following to \citet{fedus2022switch} which promotes balanced top-k selection through an auxiliary loss, aiming for an equitable workload distribution among experts.

%% file: results.tex
\section{Results and Discussion}

\begin{table}[t]
\begin{center}
\textbf{Zero-shot Results at 3B Scale}
\end{center}
\resizebox{\linewidth}{!}{
\begin{tabular}{llllccccccccc}

\toprule
\textbf{} &  \textbf{Model} & \textbf{\% Params.} & \textbf{ANLI} & \textbf{CB} &  \textbf{RTE} &  \textbf{WSC} &  \textbf{WIC} &  \textbf{Copa} &  \textbf{WNG} &  \textbf{HS}  &  \textbf{Average}  \\
\midrule

\multirow{2}{*}{\textit{Full-FT}} & T0-3B \citep{sanh2022multitask} & 100\% & 33.46  &50.0 & 64.08 & 64.42  & 50.39 & 74.92 &  50.51  & 27.51   & 51.91 \\
& T0-3B (our replication) & 100\% & 41.08 & 80.36 & 76.17 & 53.37 & 53.92 & 88.94 & 57.46 & 29.19 & 60.06 \\

\noalign{\smallskip} 
\hdashline 
\noalign{\smallskip}

\multirow{2}{*}{\textit{PEFT}} & $\text{(IA)}^3$ & 0.018\% & 34.08 &50.0  & 66.43 & 56.25 & 55.41 & 79.08 & 52.09 & 29.91 & 52.90 \\

& LORA & 0.3\% & 37.5 & 75.57 & 73.53 & 61.02 & 51.25 & 83.6 & 54.33 & 25.32  & 57.51 \\

\noalign{\smallskip} 
\hdashline 
\noalign{\smallskip}

\multirow{5}{*}{\textit{Our Method}}

& MoV-10 & 0.32\% & 38.92 & 75.0 & 78.88 & 62.5 & 52.19 & 85.77 & 55.96 & 30.24 & 59.93 \\
& MoV-30 & 0.68\% & 38.7 & 78.57 & 80.87 & 63.46 & 51.1 & 87.25 & 56.27 & 28.63 & 60.61 \\
& MoV-60 & 1.22\% & 38.83 & 76.79 & 74.55 & 60.1 & 52.66 & 89.79 & 55.49 & 30.47 & 59.83 \\
& MoLORA-10 & 3.18\% & 38.5 & 78.57 & 78.16 & 63.46 & 50.86 & 86.5 & 55.41 & 26.72 & 59.77 \\
& MoLORA-15 & 4.69\% & 40.0 & 80.36 & 80.51 & 62.98 & 50.86 & 89.0 & 55.33 & 27.3 & 60.79 \\
\bottomrule
\end{tabular}}
\vspace{0.1cm}
\caption{Average median results on unseen tasks for full model fine-tuning (T0), parameter-efficient fine-tune methods ($\text{(IA)}^3$ and LORA) and our mixture of parameter-efficient experts (MoV and MoLORA), using T5-3B base model \citep{raffel2020exploring}. Note that our replication of T0 performs significantly higher than the original T0 confirming previous work \citep{phang2023hypertuning,ivison2023hint}.}
\label{tab:main-results}
\end{table}

\textbf{Parameter efficient MoEs vs PEFTs} ~ \textit{How does our MoE recipe compare to a single expert PEFT?} Table \ref{tab:main-results} compares zero-shot performance of PEFTs methods ($\text{(IA)}^3$ and LORA), and our variants of parameter-efficient MoE (MoV and MoLORA), using T5-3B as the base model. We observe that our MoE variants (MoV and MoLORA) present a significant performance boost over the standard $\text{(IA)}^3$ vectors and LORA adapters.  

MoV using 30 experts achieves a 14.57\% performance improvement compared to its dense counterpart $\text{(IA)}^3$. This improvement is consistent across all unseen tasks and is achieved at a marginal increase in the number of updated parameters -- only an additional 0.018\% parameters per expert. 
In the context of LORA, our MoLORA equipped with 15 experts, achieves an average median score increase of 5.70\%. This improvement is notably less significant when compared to MoV. We attribute this disparity to the difference in updated parameter count in LORA adapters and $\text{(IA)}^3$ vectors (0.3\% vs 0.018\%). Overall, learning a mixture for both MoV and MoLORA as opposed to a single dense model leads to notable gains in zero-shot performance. 

\textbf{MoV outperforms MoLORA given same parameter budget} ~ Between our methods, MoV achieves a better performance-parameter cost trade-off at 3B parameters base model. As shown in the left plot in figure \ref{fig:model_params} MoV with 30 experts, only updating 0.68\% of all parameters, achieves nearly the same performance as MoLORA with 15 experts that updates 4.69\% of parameters. This shows the effectiveness of our MoE approaches even with tiny experts at a large base model scale. 

\textbf{Parameter efficient MoEs vs full fine-tuning} ~ \textit{How does MoE compare to updating all parameters during finetuning?} As shown in Table \ref{tab:main-results} when compared to fully fine-tuned T0-3B, our proposed methods, MoV and MoLORA both with 10 experts, are on par with full fine-tuning. This is impressive as MoV-10 only updates 0.32\% of all model parameters.
Furthermore, when increasing the number of experts from 10 to 15 and 30 for MoV and MoLORA respectively, our both methods outperform the full fine-tuning by a small margin.   

\label{sec:res_analysis}

\begin{figure}[t]
\centering
     \includegraphics[width=0.95\linewidth]{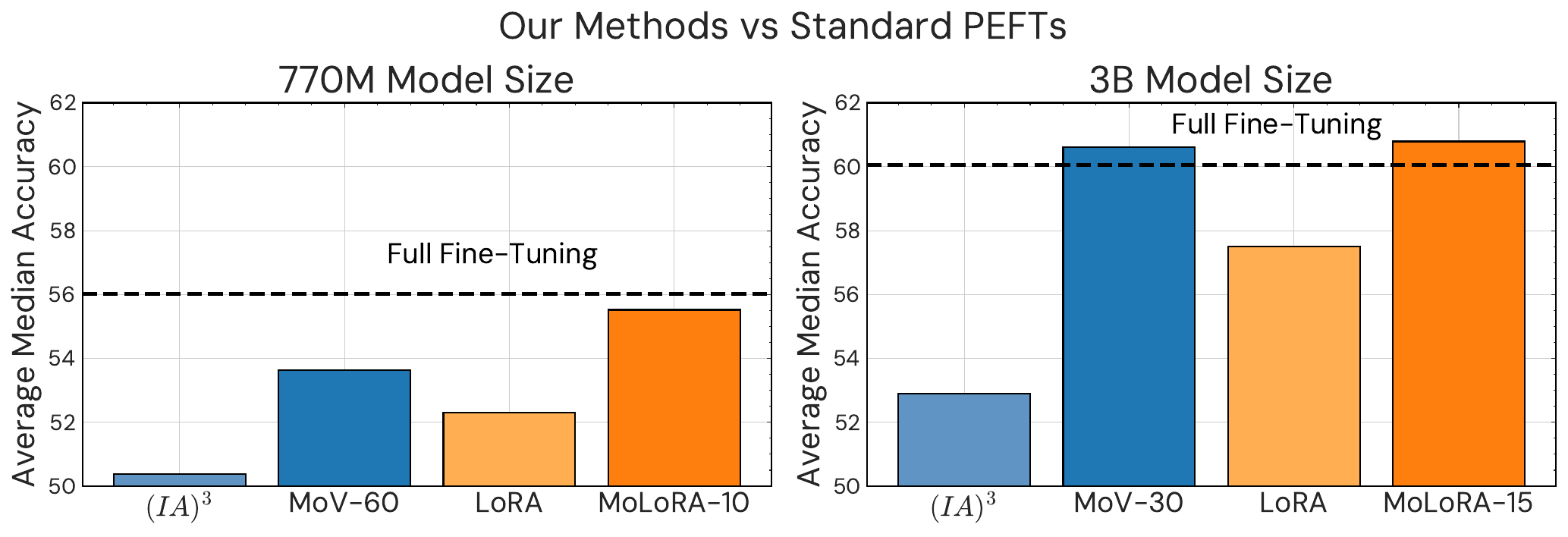}
     \caption{Comparison of the top-performing variants from our proposed mixture of PEFT experts versus their dense counterparts across T5-Large (\emph{Left}) and T5-XL (\emph{Right}).}
    \label{fig:summary-plot}  
\end{figure}

\subsection{How do parameter-efficient MoEs scale with base model size?}

Figure \ref{fig:model_params} (right) shows the scaling characteristic of MoV with 60 experts compared with $\text{(IA)}^3$ and full fine-tuning for 770M, 3B and 11B parameters base models. We find that across all model sizes we evaluate, our parameter-efficient MoEs consistently maintain higher performance compared to standard PEFTs and achieve comparable results with full fine-tuning.

\textbf{MoV benefits from scaling} At all model sizes, MoV-60 significantly outperforms standard $\text{(IA)}^3$. It is also far closer in performance to full fine-tuning than a single expert. For example, at 770M parameters, there is a 12.34\% performance gap between $\text{(IA)}^3$ and full fine-tuning vs 5.56\% for MoV-60. As the base model scales up, MoV becomes more competitive with full fine-tuning. For 3B and 11B parameter models, MoV-60 achieves performance approximately on par with the full fine-tuning, despite updating less than 1.3\% of the total parameters. 


\textbf{MoLORA outperforms MoV in smaller model size regimes} As discussed in the main results, at larger model sizes MoV achieves a better performance-parameter efficiency trade-off compared to MoLORA. Conversely, at the 770M scale, MoLORA with 10 experts that updates 3.18\% of total parameters, performs better compared to MoV-60 and nearly matches the performance of full fine-tuning (Figure \ref{fig:summary-plot}). Finally, similar to MoV, MoLORA archives higher performance than LORA at both 770M and 3B scales.

\subsection{How does the number of experts impact the downstream performance?}
\label{sec:num-experts}

\begin{figure*}[t]
    \centering
        \includegraphics[width=1.0\linewidth]{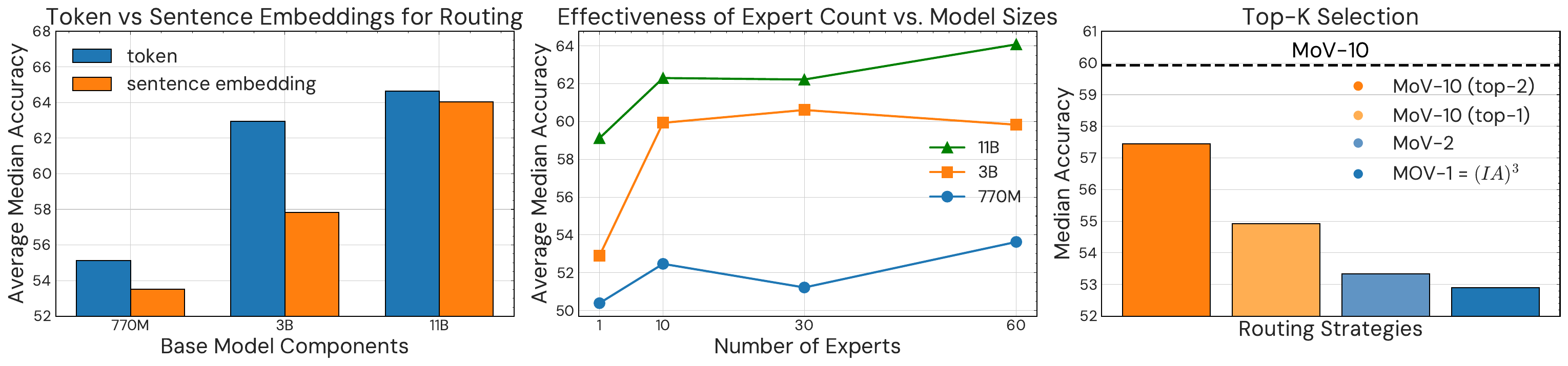} 
    \caption{\emph{Left:} Zero-shot performance of passing embedding of the token sequence to the router vs. passing tokens to the router. \emph{Middle:} Zero-shot performance across T5 model sizes (Large, XL, XXL) as the number of experts increases. \emph{Right:} The effectiveness of activating top-k experts.}
    \label{fig:performance_embedd_expert_router}
\end{figure*}

The center plot of Figure \ref{fig:performance_embedd_expert_router} shows the performance of MoV with different numbers of experts at all model sizes. We find that increasing the number of experts generally improves unseen task performance. However, this improvement is contingent upon the specific number of experts and the base model size. 
For both 770M and 11B parameter base models, our MoV method achieves its best performance by using 60 experts. To illustrate, when number of experts is increased from 10 to 60, the average median accuracy improves from 52.47 to 53.63 for the 770M model and from 62.3 to 64.08 for the 11B model. However, for the 3B model, using just 30 experts, updating 0.68\% of the parameters, reaches peak accuracy with a score of 60.61 at this scale, as performance stagnates when 60 experts are used.

This trend of performance improvement by scaling more experts is further corroborated in the context of MoLORA; when scaling experts from sets of (5, 10, 15), there was a corresponding elevation in the average median score, registering at 58.6, 59.77, and 60.79, respectively.

\subsection{What is the best routing strategy in parameter-efficient MoEs?}

In Figure \ref{fig:performance_embedd_expert_router}, the rightmost plot shows the overall unseen task performance when using different routing strategies for MoV. Specifically, we compare the \textit{soft merging} of 10 experts (dashed line) with discrete top-2 and top-1 routing. We observe that soft merging significantly outperforms discrete routing in the MoV-10 setting. Specifically, for discrete routing with top-k experts, where k is 1 and 2, the MoE achieves an average median accuracy of 54.92 and 57.45 respectively. In contrast, using the soft merging approach, where all experts are activated, we observe an accuracy of 59.93. 

Furthermore, to understand if we recover the performance loss of top-k routing by using load balancing, we integrated the loss balancing following to \cite{fedus2022switch}. However, we find that the top-k selection of $k=2$ with load balancing loss leads to a further decrease in performance 1.5 average median score. 

Together, these results show that in extremely parameter-efficient MoE settings, soft merging enables superior performance. Note that top-2 and top-1 routing strategies (among 10 experts) perform better than MoV with only 2 experts and a single expert $\text{(IA)}^3$ respectively, showing that soft merging performs better when a larger number of experts are used. 

\subsection{Does a pronounced task information in routing lead to higher performance?}
\label{sec:token_v_embedding}

To understand the effects of a pronounced inductive bias towards task representations in our MoE framework, we compare using sentence embeddings of instructions with token embeddings for the routing input. These sentence embeddings are obtained offline using an external sentence embedding model. Here, we aim to evaluate how pronounced task information affects the router's decision and the subsequent generalization capabilities of the model in downstream tasks. Figure \ref{fig:performance_embedd_expert_router} leftmost plot shows performances of token routing and sentence routing at all model sizes. We find that the token routing exhibits superior performance with 3.03\%, 8.86\%, and 0.94\% improvement for 770M, 3B, and 11B base model sizes respectively. These results suggest that a higher degree of inductive bias for task datasets is not necessarily beneficial as our approaches can acquire a diverse set of task knowledge directly from the hidden representations of tokens. Furthermore, token routing enables the use of learned experts and routing layers without any prior task information for unseen tasks.

\begin{figure*}[t]
    \centering
    \includegraphics[width=0.9\linewidth]{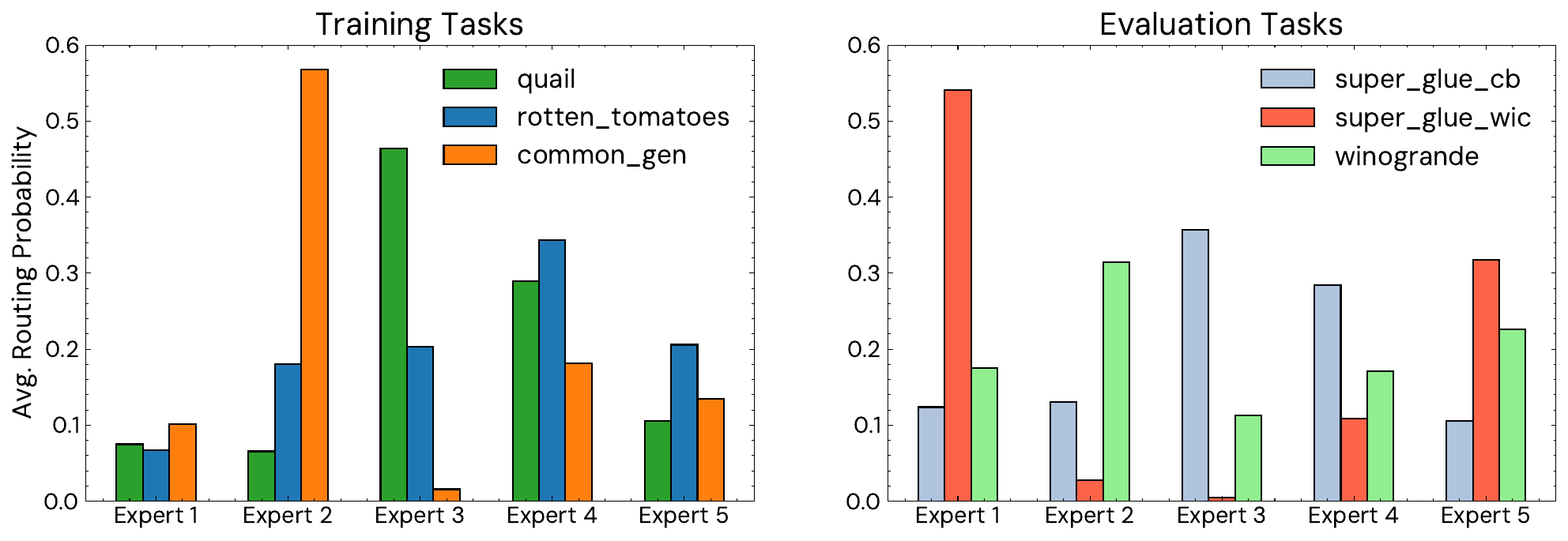}
    \caption{Mean expert routing probabilities for intermediates activations at the last feedforward layer. Values are averaged across tokens and batch. Experts are weighted differently in soft merging depending on the task. \emph{Left:} Measured on tasks seen during training. \emph{Right:} Measured on unseen evaluation tasks.}
    \label{fig:expert-dists}
\end{figure*}

\subsection{Do experts specialize in diverse knowledge across different tasks?}
\label{sec:expert-distribution}

To understand how expert routing differs for different tasks, we take a closer look at how experts are activated for a variety of tasks. Figure \ref{fig:expert-dists} shows the mean expert probabilities for MoV with 5 experts that are located in feed-forward layers in the last decoder block at 770M parameter T5 model. We selected the last decoder block as it has been shown deeper layers learn more task-specific information \citep{rogers2020primer}. We plot the mean routing probabilities for both training tasks and evaluation tasks that are unseen during training, to understand cross-task generalization through the lens of experts if skills learned at training time generalize to unseen tasks at evaluation time. 
Intuitively, if experts have indeed learned different \textit{skills}, we expect that they contribute in different degrees to tasks that are different in nature. The amount of contribution is directly reflected in the routing probability of each expert since we use soft merging i.e. summation of expert vectors weighted by the routing probability as described in Figure \ref{fig:mov-architecture}. As such, the mean routing probabilities plotted in Figure \ref{fig:expert-dists} provide an overall picture of the contribution of each expert, depending on the downstream task.

\textbf{Specialization across unseen vs seen tasks} As depicted in Figure \ref{fig:expert-dists}, both evaluation and training tasks lead to the activation of experts at different magnitudes. For example, both \texttt{quail} and \texttt{super\_glue\_cb} activate Expert 3 the most out of the 5 experts, followed by Expert 4 but are different both in terms of the relative contribution of each expert and the ordering of the remaining 3 experts based on routing probability. A similar pattern can be observed for \texttt{common\_gen} \& \texttt{winogrande} as they both activate Expert 2 the most but are otherwise different. Overall, the fact that routing specialization seems to occur \emph{regardless} of whether the downstream task was trained on, suggests that expert specialization is inherent and transferable from seen tasks to unseen tasks.

\subsection{Hyper-parameters Sensitivity} 
Given the widely documented sensitivity of MoE-style architecture to hyperparameters \citep{fedus2022switch, shazeer2017outrageously}, we ran extensive ablation studies to uncover the idiosyncrasy of PEFT methods in the context of MoE. We experimented with batch sizes of 32, 128, 256, and 2048 and we found that the larger the batch size, the more likely our MoEs to collapse to a single expert. Our empirical finding resonates with \cite{shen2023mixtureofexperts} which also finds that a small batch is necessary for stable training. For instance, by experimenting with a batch size of 2048 and evaluating every 5K steps up to 20K, we observed that the performance of our parameter-efficient MoEs deteriorates after 5K steps, converging to performance levels akin to their dense counterparts. Additionally, we experimented with varying learning rates from $3e^{-3}$ to $6e^{-4}$ where we discovered for our methods, a smaller learning rate of $3e^{-4}$ leads to higher performance relative to their dense PEFT counterpart and full fine-tuning. Smaller learning rates stabilize training in parameter-efficient experts by preventing rapid, imbalanced updates that can suppress diversity and lead to suboptimal solutions.

%% file: related_work.tex
\section{Related Work}
\label{sec:related}

\textbf{Mixture-of-Experts} ~ The Mixture-of-Experts (MoE) has been investigated thoroughly in Natural Language Processing \citep{lou2022crosstoken,mustafa2022multimodal, shazeer2017outrageously,lepikhin2020gshard, fedus2022switch, du2022glam,zoph2022stmoe,clark2022unified, zhou2022mixtureofexperts, komatsuzaki2023sparse,kudugunta2021distillation, zuo2022taming} as an effective way of increasing the model's capacity in parameter size where certain parts of the model are activated while computation is kept the same or close to its dense counterpart. In the context of MoE, there is a body of work focusing on improving the routing \citep{hazimeh2021dselectk, pmlr-v139-lewis21a, roller2021hash, zhou2022mixtureofexperts} including random routing \citep{zuo2022taming} activating all expert through weighted average \citep{eigen2014learning} to sparsely select a single or $k$ experts \citep{fedus2022switch, du2022glam}. MoE has also been invested in multi-task settings including multilingual neural machine translation\citep{hazimeh2021dselectk, kudugunta2021distillation}. Unlike these studies, our research addresses MoE by scaling both the volume of data and the number of tasks, aiming to mitigate the instability inherent in training the MoE models. But our primary emphasis remains on achieving efficient fine-tuning. Recently, \cite{shen2023mixtureofexperts} highlighted how instruction fine-tuning with scaled tasks can counteract the generalization challenges tied to MoE models. In distinction from this, our study scrutinizes the efficacy of instruction fine-tuning in the MoE domain, specifically concentrating on a unique ensemble of the PEFT components, considering the memory cost of the traditional MoE can be prohibitive for many practitioners. Similar to the aforementioned work, \cite{ye2022eliciting} utilized MoE in a multi-task context, employing BART \cite{lewis2019bart} as their pre-trained model. However, they limited their experimental scope to a smaller scale and used replicas of each transformer layer as experts, simply multiplying the model by the number of experts. Our work, on the other hand, presents an extreme parameter efficiency with small experts at a large scale up to 11B parameter base model. 


\textbf{Instruction Tuning}
Instruction tuning, as elucidated in \citep{sanh2022multitask, wei2022finetuned, mishra2022crosstask}, is a technique where a language model is fine-tuned over a collection of tasks using paired prompts and responses. The primary goal of this technique is to enable the model to predict responses accurately based on the provided prompts, thereby augmenting its ability to understand and execute instructions effectively. The method has gained considerable attention due to its pronounced success in enhancing zero-shot performance on tasks to which the model has not been previously exposed. Additionally, instruction tuning has led to breakthroughs such as Chain of Thought Prompting \citep{wei2023chainofthought} where a breakdown of complex problems into smaller steps to produce intermediate reasoning along with the final solution, PaLM \citep{chowdhery2022palm}, FLAN \citep{wei2022finetuned}. 
In our work, we explore the use of instruction fine-tuning with the intention of harnessing its benefits that enable the model to learn from a diverse set of inputs where the mixture of expert style models suits well, for enhanced evaluation performance on unseen tasks. Our objective remains to optimize computational efficiency without compromising zero-shot performance.

\textbf{Parameter-Efficient Fine-tuning}. 
\cite{houlsby2019parameterefficient} established "adapters" in the NLP domain to fine-tune BERT. There are many variants of adapters with different design choices \citep{bapna2019simple,pfeiffer2021adapterfusion}. \cite{ li2021prefixtuning} proposed updating soft prompts concatenated to embeddings or layer outputs instead of adapters. \cite{zaken2022bitfit} show that just updating only a small
subset of parameters during fine-tuning (e.g. just biases) is very effective. \cite{hu2021lora} proposed LORA based on low-rank decomposition matrices of transformer layers. They show superior performance with a smaller parameter budget and no inference cost as LORA parameters can be applied offline to the baseline model. \cite{liu2022fewshot} proposed $(IA)^3$, task-specific vectors to modify attention activation. Instead of using feedforward layers inserted in transformer layers as adapters, they learn vectors to update (by broadcast multiplication) key, value, and linear layer weight matrices. Unlike the other PEFT methods, $\text{(IA)}^3$ does not induce any additional inference cost and enables mix-batches (from different datasets). The multiplicative nature of the $\text{(IA)}^3$ creates an interesting opportunity for the mixture-of-expert type of modeling without parallelization overhead. \cite{chen2023parameterefficient} experiment with different design spaces (essentially a hyperparameter search) for PEFT. They suggest four phases: 1) grouping layers into different sets; 2) adding trainable parameters towards each group; 3) deciding which group should be trained; 4) assigning groups with different training strategies. Their finding is that different architectures have different best settings. We have chosen $(IA)^3$ and LORA as our PEFT components because they offer an optimal balance between performance and parameter efficiency \citep{mahabadi2021parameterefficient, liu2022fewshot}.

Several studies have explored PEFT in the context of MoE or in a similar fashion, albeit with certain distinctions. For instance, \cite{wang2022adamix} focused on single-task fine-tuning employing a mixture of adapters for BERT$_{base}$ with 110M parameters \citep{devlin2019bert} and $RoBERTa_{large}$ with 355M parameters \citep{liu2019roberta}, incorporating random routing, and adopting a few-shot evaluation. In divergence from this, our work centers on instruction-tuning with multiple tasks present during fine-tuning. We underscore the efficacy of this approach by rigorously testing up to 11B parameter text-to-text model \cite{raffel2020exploring}, implementing token routing, and strictly emphasizing evaluation on a set of unseen (held-out) tasks to underscore the potential of instruction tuning. In another work, \cite{ponti2022combining} introduced Polytropon, which involves learning adapters (termed as 'skills') specific to each task and employing a task-skills binary matrix to determine the skill set associated with each task. In their method, input examples dictate the selection of adapters. These adapters are then aggregated, and the resultant single adapter is integrated into the overall architecture. Extending upon the Polytropon framework, \cite{caccia2023multihead} implemented a distinct skill set for every layer in their variant named Polytropon-S. They introduce a deterministic routing function, delve into supplementary inductive biases, show effectiveness up to 3B models, and they don't employ MoE style architecture. Our research presents a departure from these two studies. Specifically, our primary experimental setup employs MoEs that do not require any specific task identifier during fine-tuning by the use of their token routing strategy. In this way, we can evaluate our instruction-tuned MoEs on unseen tasks without any further task-specific few-shot fine-tuning. We showed the scaling property of our MoEs in this setting by fine-tuning models up to 11B parameters.

%% file: appendix.tex
\setcounter{section}{0}
\renewcommand{\thesection}{\Alph{section}}

\section{Full Experimental Results}

\label{sec:appendix}

\subsection{Zero-Shot Evaluation for P3 dataset}

In our study, we conducted a comprehensive evaluation of the variants of our proposed methods in comparison to our established baselines. This evaluation encompassed various sizes of the T5 model, specifically 770M, 3B, and 11B. Both mean and median scores were reported for every evaluation set derived from the P3 dataset, which covers a range of tasks. For further details and a more in-depth exploration, please refer to the following URL: \url{https://huggingface.co/datasets/bigscience/P3}.

\begin{table}[h]
\begin{center}
\textbf{T5-Large (770M)}
\end{center}
\label{tab:t5_large_results}

\resizebox{\linewidth}{!}{
\begin{tabular}{llllccccccccc}

\toprule
\textbf{} &  \textbf{Model} & \textbf{\% Params.} & \textbf{Metric} & \textbf{ANLI} & \textbf{CB} &  \textbf{RTE} &  \textbf{WSC} &  \textbf{WIC} &  \textbf{Copa} &  \textbf{WNG} &  \textbf{HS}  &  \textbf{Average}  \\
\midrule
\multirow{2}{*}{\textit{Full-FT}} & \multirow{2}{*}{T0-770M (ours)} & 100\% & median & 35.6 & 71.43 & 75.63 & 57.21 & 51.41 & 77.0 & 53.04 & 26.78 & 56.01 \\
& & & mean & 35.57 & 57.74 & 75.88 & 52.31 & 52.52 & 74.6 & 52.93 & 26.74 & 53.54 \\

\noalign{\smallskip} 
\hdashline 
\noalign{\smallskip}

\multirow{4}{*}{\textit{PEFT}} & $\text{(IA)}^3$ & 0.036\% & median & 33.5 & 42.86 & 67.87 & 62.02 & 52.35 & 67.0 & 51.22 & 26.33 & 50.39 \\ 
& & & mean & 33.27 & 45.12 & 67.08 & 58.17 & 52.74 & 66.63 & 51.35 & 26.32 & 50.09 \\
& LoRA & 0.497\% & median & 35.0 & 55.36 & 57.4 & 63.46 & 50.24 & 77.0 & 53.28 & 26.67 & 52.3 \\
& & & mean & 35.26 & 51.67 & 59.35 & 62.98 & 50.66 & 76.5 & 52.41 & 27.24 & 52.0 \\

\noalign{\smallskip} 
\hdashline 
\noalign{\smallskip}

\multirow{10}{*}{\textit{Our Method}} & MOV-5 & 0.27\% & median & 33.6 & 41.07 & 71.48 & 61.54 & 50.86 & 76.5 & 51.46 & 26.02 & 51.57 \\
& & & mean & 33.51 & 42.62 & 71.26 & 60.96 & 51.14 & 73.8 & 51.55 & 26.01 & 51.36 \\
& MoV-10 & 0.55\% & median & 33.9 & 42.86 & 74.19 & 62.5 & 50.31 & 77.0 & 52.64 & 26.34 & 52.47 \\
& & & mean & 33.68 & 42.38 & 74.51 & 59.23 & 50.74 & 74.82 & 52.2 & 26.72 & 51.78 \\
& MoV-20 & 1.10\% & median & 33.7 & 41.07 & 73.83 & 63.46 & 50.94 & 75.46 & 51.14 & 25.48 & 51.89 \\
& & & mean & 33.98 & 45.12 & 73.36 & 59.13 & 51.33 & 73.47 & 51.3 & 25.45 & 51.64 \\
& MoV-30 & 1.66\% & median & 33.75 & 41.07 & 72.92 & 55.77 & 51.25 & 77.0 & 51.46 & 26.55 & 51.22 \\
& & & mean & 33.81 & 44.88 & 72.56 & 56.15 & 51.29 & 77.43 & 51.81 & 26.52 & 51.81 \\
& MoV-60 & 3.32\% & median & 34.0 & 53.57 & 75.81 & 57.69 & 50.55 & 77.96 & 53.12 & 26.33 & 53.63 \\
& & & mean & 34.24 & 52.26 & 75.02 & 58.37 & 50.78 & 77.06 & 52.87 & 26.74 & 53.42 \\
& MoLoRA-10 & 5.60\% & median & 33.2 & 67.86 & 68.41 & 64.9 & 50.39 & 80.0 & 52.64 & 52.64 & 55.52 \\
& & & mean & 33.37 & 56.31 & 68.88 & 63.37 & 51.55 & 79.35 & 52.31 & 52.31 & 53.99 \\

\bottomrule
\end{tabular}}
\vspace{0.1cm}
\caption{Zero-shot evaluation of the 770M parameter model across all unseen tasks, comparing different numbers of experts for both MoV and MoLoRA.}
\label{tab:results}
\end{table}

\begin{table}[h]
\begin{center}
\textbf{T5-XL (3B)}
\end{center}
\label{tab:t5_xl_results}

\resizebox{\linewidth}{!}{
\begin{tabular}{llllccccccccc}

\toprule
\textbf{} &  \textbf{Model} & \textbf{\% Params.} & \textbf{Metric} & \textbf{ANLI} & \textbf{CB} &  \textbf{RTE} &  \textbf{WSC} &  \textbf{WIC} &  \textbf{Copa} &  \textbf{WNG} &  \textbf{HS}  &  \textbf{Average}  \\
\midrule

\multirow{4}{*}{\textit{Full-FT}} & \multirow{2}{*}{T0-3B \citep{sanh2022multitask}} & 100\% & median & 33.46  &50.0 & 64.08 & 64.42  & 50.39 & 74.92 &  50.51  & 27.51   & 51.91 \\
& & & mean & 33.42 & 45.36 & 64.55 & 65.10 & 50.69 & 72.40 & 50.97 & 27.29 & 51.22 \\
& \multirow{2}{*}{T0-3B (our replication)} & 100\% & median & 41.08 & 80.36 & 76.17 & 53.37 & 53.92 & 88.94 & 57.46 & 29.19 & 60.06 \\
& & & mean & 40.73 & 74.52 & 76.82 & 52.21 & 53.84 & 88.99 & 56.83 & 29.2 & 59.14 \\

\noalign{\smallskip} 
\hdashline 
\noalign{\smallskip}

\multirow{4}{*}{\textit{PEFT}} & $\text{(IA)}^3$ & 0.018\% & median & 34.08 &50.0  & 66.43 & 56.25 & 55.41 & 79.08 & 52.09 & 29.91 & 52.90 \\ 
& & & mean & 34.56 & 51.07 & 68.38 &  54.9 & 55.61 & 78.23 & 52.14 & 28.97 & 52.98 \\

& LoRA & 0.3\% & median & 37.5 & 75.57 & 73.53 & 61.02 & 51.25 & 83.6 & 54.33 & 25.32  & 57.51 \\
& & & mean & 37.85 & 66.9 & 77.04 & 56.73 & 52.29 & 82.83 & 55.64 & 26.79 & 57.01 \\

\noalign{\smallskip} 
\hdashline 
\noalign{\smallskip}

\multirow{24}{*}{\textit{Our Method}} 
& MoV-2 & 0.18\% & median & 34.7 & 46.43 & 66.06 & 56.25 & 54.86 & 85.42 & 53.75 & 29.25 & 53.34 \\
& & & mean & 35.14 & 50.36 & 69.31 & 56.15 & 54.4 & 83.79 & 53.69 & 28.47 & 53.91 \\
& MoV-5 & 0.23\% & median & 37.1 & 76.79 & 78.16 & 57.69 & 52.27 & 86.77 & 53.99 & 29.31 & 59.01 \\
& & & mean & 37.66 & 62.14 & 78.3 & 58.46 & 53.54 & 86.52 & 54.54 & 28.3 & 57.43 \\
& MoV-10 & 0.32\% & median & 38.92 & 75.0 & 78.88 & 62.5 & 52.19 & 85.77 & 55.96 & 30.24 & 59.93 \\
& & & mean & 38.83 & 63.45 & 79.49 & 60.19 & 53.04 & 86.41 & 56.27 & 29.11 & 58.35 \\
& MoV-20 & 0.50\% & median & 39.2 & 75.0 & 76.71 & 57.69 & 53.45 & 89.0 & 55.64 & 30.89 & 59.7 \\
& & & mean & 39.25 & 64.05 & 76.53 & 56.63 & 53.45 & 86.93 & 56.24 & 29.36 & 57.81 \\
& MoV-30 & 0.68\% & median & 38.7 & 78.57 & 80.87 & 63.46 & 51.1 & 87.25 & 56.27 & 28.63 & 60.61 \\
& & & mean & 38.9 & 67.5 & 81.23 & 59.9 & 52.43 & 86.28 & 56.39 & 27.57 & 58.77 \\
& MoV-60 & 1.22\% & median & 38.83 & 76.79 & 74.55 & 60.1 & 52.66 & 89.79 & 55.49 & 30.47 & 59.83 \\
& & & mean & 38.97 & 63.93 & 75.38 & 57.79 & 53.5 & 86.04 & 55.88 & 29.28 & 57.59 \\
& MoV-10 (top-1) & 0.32\% & median & 33.9 & 75.0 & 71.12 & 61.06 & 50.71 & 70.0 & 51.7 & 25.89 & 54.92 \\
& & & mean & 34.31 & 60.6 & 71.41 & 58.94 & 51.24 & 68.39 & 51.79 & 25.98 & 52.82 \\
& MoV-10 (top-2) & 0.32\% & median & 38.7 & 82.14 & 75.63 & 48.08 & 53.68 & 79.88 & 54.14 & 27.37 & 57.45 \\
& & & mean & 38.89 & 69.76 & 74.95 & 47.69 & 53.51 & 79.89 & 53.83 & 26.91 & 55.67 \\
& MoLORA-2 & 0.75\% & median & 39.2 & 82.14 & 80.32 & 62.5 & 50.39 & 80.58 & 57.38 & 28.47 & 60.12 \\
& & & mean & 38.86 & 65.71 & 80.0 & 60.0 & 50.8 & 82.17 & 56.51 & 28.03 & 57.76 \\
& MoLORA-5 & 1.66\% & median & 36.75 & 71.43 & 79.96 & 56.25 & 55.17 & 85.81 & 55.8 & 27.63 & 58.6 \\
& & & mean & 37.52 & 62.14 & 80.22 & 52.6 & 55.34 & 84.05 & 56.04 & 26.62 & 56.82 \\
& MoLORA-10 & 3.18\% & median & 38.5 & 78.57 & 78.16 & 63.46 & 50.86 & 86.5 & 55.41 & 26.72 & 59.77 \\
& & & mean & 38.49 & 66.43 & 77.44 & 59.9 & 51.63 & 84.96 & 56.1 & 26.7 & 57.71 \\
& MoLORA-15 & 4.69\% & median & 40.0 & 80.36 & 80.51 & 62.98 & 50.86 & 89.0 & 55.33 & 27.3 & 60.79 \\
& & & mean & 39.73 & 69.52 & 80.97 & 60.67 & 51.54 & 86.5 & 55.03 & 27.25 & 58.9 \\

\bottomrule
\end{tabular}
}
\caption{In our most comprehensive experimental setup, we conducted a zero-shot evaluation across all unseen tasks using a 3B parameter model. We compared varying numbers of experts for both MoV and MoLoRA and experimented with a top-k selection routing strategy}
\end{table}

\begin{table}[h]
\begin{center}
\textbf{T5-XXL (11B)}
\end{center}
\label{tab:t5_xxl_results}

\resizebox{\linewidth}{!}{
\begin{tabular}{llllccccccccc}

\toprule
\textbf{} &  \textbf{Model} & \textbf{\% Params.} & \textbf{Metric} & \textbf{ANLI} & \textbf{CB} &  \textbf{RTE} &  \textbf{WSC} &  \textbf{WIC} &  \textbf{Copa} &  \textbf{WNG} &  \textbf{HS}  &  \textbf{Average}  \\
\midrule

\multirow{4}{*}{\textit{Full-FT}} 
& T0-11B \citep{sanh2022multitask} & 100\% & median & 42.17 & 78.57 & 81.23 & 64.42 & 57.21 & 90.79 & 60.46 & 33.65 & \textbf{63.56} \\
& & & mean & 41.16 & 70.12 & 80.83 & 61.45 & 56.58 & 90.02 & 59.94 & 33.58 & \textbf{61.70} \\
& T0-11B (our replication) & 100\% & median & 47.1 & 80.36 & 81.41 & 60.1 & 56.27 & 96.08 & 67.32 & 31.61 & \textbf{65.03} \\
& & & mean & 45.83 & 72.62 & 81.52 & 58.17 & 56.66 & 96.0 & 66.77 & 30.95 & \textbf{63.57} \\

\noalign{\smallskip} 
\hdashline 
\noalign{\smallskip}

\multirow{2}{*}{\textit{PEFT}} 
& $\text{(IA)}^3$ & 0.0098\% & median & 42.3 & 73.21 & 75.99 & 58.65 & 52.04 & 86.27 & 54.3 & 30.27 & \textbf{59.12} \\
& & & mean & 42.1 & 63.27 & 75.31 & 55.49 & 52.27 & 85.74 & 55.06 & 30.09 & \textbf{57.41} \\

\noalign{\smallskip} 
\hdashline 
\noalign{\smallskip}

\multirow{8}{*}{\textit{Our Method}} 
& MoV-10 & 0.143\% &  median & 45.83 & 76.79 & 78.52 & 53.85 & 51.88 & 94.23 & 63.77 & 33.5 & \textbf{62.3} \\
& & & mean & 44.73 & 70.12 & 78.88 & 54.23 & 53.26 & 93.64 & 63.57 & 33.59 & \textbf{61.5} \\

& MoV-20 & 0.287\% & median & 44.58 & 76.79 & 73.83 & 55.77 & 52.98 & 95.0 & 62.27 & 32.92 & \textbf{61.77} \\
& & & mean & 43.54 & 69.17 & 74.4 & 52.88 & 54.5 & 93.93 & 62.95 & 32.85 & \textbf{60.53} \\

& MoV-30 & 0.431\% & median & 43.6 & 76.79 & 77.62 & 56.73 & 53.84 & 93.62 & 64.25 & 31.34 & \textbf{62.22} \\
& & & mean & 43.32 & 69.29 & 77.22 & 53.56 & 56.03 & 93.65 & 63.52 & 31.32 & \textbf{60.99} \\

& MoV-60 & 0.862\% & median & 45.17 & 75.0 & 83.03 & 60.1 & 53.68 & 95.42 & 65.82 & 34.38 & \textbf{64.08} \\
& & & mean & 43.9 & 69.88 & 83.07 & 56.54 & 54.51 & 94.01 & 64.56 & 34.17 & \textbf{62.58} \\

\bottomrule
\end{tabular}}
\vspace{0.1cm}
\caption{We evaluated the largest available model size from the original T5 pre-trained checkpoint, T5-XXL with 11B parameters, to demonstrate the efficacy of our proposed mixture of PEFT experts at this scale.}
\label{tab:results}
\end{table}

\clearpage

\subsection{Token vs. Sentence Embeddings for Routing}

We present the mean and median results for our routing strategies. Specifically, we assessed performance by either passing tokens directly to the router or by passing sentence embeddings. Our findings indicate that, particularly for the T5-XL(3B) model, token routing consistently yields better performance in terms of both mean and median values. The Anli dataset is excluded from our embedding dataset.

\begin{table}[h]
\begin{center}
\textbf{MoV -- Token vs. Sentence Embedding}
\end{center}
\label{tab:prod_results}

\resizebox{\linewidth}{!}{\begin{tabular}{lllccccccccc}

\toprule
\textbf{} &  \textbf{Model} & \textbf{Metric}  & \textbf{CB} &  \textbf{RTE} &  \textbf{WSC} &  \textbf{WIC} &  \textbf{Copa} &  \textbf{WNG} &  \textbf{HS}  &  \textbf{Average}  \\
\midrule

& MoV-10 (Token) - 770M & median & 42.86 & 74.19 & 62.5 & 52.64 & 52.64 & 77.0 & 26.34 & 55.12 \\
& & mean & 42.38 & 74.51 & 59.23 & 52.2 & 52.2 & 74.82 & 26.72 & 54.37 \\

& MoV-10 (Embedding) - 770M & median & 48.21 & 67.15 & 62.98 & 51.8 & 50.99 & 67.0 & 26.38 & 53.5 \\
& & mean & 51.67 & 67.29 & 58.37 & 51.79 & 50.99 & 65.8 & 26.57 & 53.21 \\

& MoV-10 (Token) - 3B & median & 75.0 & 78.8 & 62.5 & 52.19 & 55.96 & 85.77 & 30.24 & 62.94 \\
& & mean & 63.45 & 79.49 & 60.19 & 53.04 & 56.27 & 86.41 & 29.11 & 61.14 \\

& MoV-10 (Embedding) - 3B & median & 57.14 & 67.15 & 61.06 & 55.33 & 52.49 & 82.5 & 29.08 & 57.82 \\
& & mean & 51.07 & 68.81 & 58.65 & 55.28 & 52.57 & 80.53 & 28.51 & 56.49 \\

& MoV-10 (Token) - 11B & median & 76.79 & 78.52 & 53.85 & 51.88 & 63.77 & 94.23 & 33.5 & 64.65 \\
& & mean & 70.12 & 78.88 & 54.23 & 53.26 & 63.57 & 93.64 & 33.59 & 63.9 \\

& MoV-10 (Embedding) - 11B & median & 75.0 & 78.7 & 57.69 & 54.0 & 57.85 & 92.0 & 33.08 & 64.05 \\
& & mean & 66.19 & 79.1 & 58.37 & 54.83 & 58.78 & 91.17 & 32.7 & 63.02 \\

\bottomrule
\end{tabular}}
\vspace{0.1cm}
\caption{The above results demonstrate the effectiveness of token routing in comparison to imposing a strong inductive bias, such as sentence embedding across various model parameters.}
\label{tab:mov_embedding_results}
\end{table}